\pdfoutput=1
\documentclass[letterpaper]{article} 

\def\year{2018}\relax
\usepackage{aaai18}  
\usepackage{times}  
\usepackage{helvet}  
\usepackage{courier}  
\usepackage{url}  
\usepackage{graphicx}  
\usepackage{balance}

\usepackage[square,sort,comma,numbers]{natbib}

\usepackage[notheorems]{rr-math}

\usepackage{nicefrac}

\usepackage[table]{xcolor}

\usepackage{xcolor}

\usepackage{oubraces}

\usepackage{verbatim}
\usepackage{wrapfig}

\usepackage{booktabs}

\newcommand{\dataName}[1]{\ensuremath{\fontsize{6}{7.5}\selectfont \mathsf{#1}}} 

\usepackage{graphicx}
\usepackage{multirow}
\usepackage{rotate}
\usepackage{subfigure}
\usepackage{epstopdf}
\usepackage{paralist}
\usepackage{url}
\usepackage{graphicx}
\usepackage{wrapfig}
\usepackage{graphicx}

\usepackage{amssymb}
\usepackage{amsmath}
\usepackage{amsfonts}

\newcommand{\todo}[1]{}

\newcommand{\incomplete}[1]{}

\usepackage{tabularx}
\usepackage{booktabs}

\usepackage{relsize}
\usepackage{algorithm}
\usepackage{algorithmicx}
\usepackage{algpseudocode}

\algblockdefx[parallel]{ParFor}{EndPar}[1][]{$\textbf{parallel for}$ #1 $\textbf{do}$}{$\textbf{end parallel}$}
\algrenewcommand{\alglinenumber}[1]{\fontsize{6.5}{7}\selectfont#1}
\algtext*{EndPar}

\algblockdefx[parallel]{parfor}{endpar}[1][]{$\textbf{parallel for}$ #1 $\textbf{do}$}{$\textbf{end parallel}$}
\algrenewcommand{\alglinenumber}[1]{\scriptsize#1:}

\usepackage{nicefrac}

\algnotext{EndIf}
\algnotext{EndProcedure}

\ifpdf
\usepackage{epstopdf}
\fi
\graphicspath{{./}{./figures/}}

\graphicspath{{graphics/}{../graphics/}{./}}

\usepackage{array}
\newcolumntype{P}[1]{>{\centering\arraybackslash}p{#1}}
\newcolumntype{M}[1]{>{\centering\arraybackslash}m{#1}}

\newtheorem{mydef}{Definition}

\newcommand{\eol}{\end{enumerate}\setlength{\itemsep}{-\parsep}}

\relax

\usepackage{wrapfig}
\usepackage{xspace}

\usepackage{csquotes}

\newcommand{\etal}{\emph{et al.}\xspace}

\newlength{\commentWidth}
\newcommand{\bspacing}{\begin{spacing}{1.4}}
\newcommand{\espacing}{\end{spacing}}

\definecolor{plotblue}{RGB}	{30,144,255}
\definecolor{plotgreen}{RGB}	{50,205,50}
\definecolor{plotred}{RGB}	{220,20,60}

\definecolor{myyellow}{RGB}{255,255,204}
\definecolor{myred}{RGB}{255,204,204}
\definecolor{myblue}{RGB}{204, 255, 255}
\definecolor{mygreen}{RGB}{204, 255, 204}

\definecolor{gray}{RGB}{150,150,150}
\definecolor{theblue}{RGB}{0,0,180}

\usepackage{textcase}
\usepackage{soul}

\newcommand*\hrulefillvar[1][0.4pt]{\leavevmode\leaders\hrule height#1\hfill\kern0pt}

\newcommand{\be}{\begin{equation}}
\newcommand{\ee}{\end{equation}}
\newcommand{\bea}{\begin{eqnarray}}
\newcommand{\eea}{\end{eqnarray}}
\newcommand{\bit}{\begin{itemize}}
\newcommand{\eit}{\end{itemize}}

\definecolor{lightgray}{rgb}{0.93,0.93,0.93}

\definecolor{lightblue}{rgb}{0.5,0.90,1.0}
\definecolor{lightgreen}{rgb}{0.5,0.92,0.5}
\definecolor{lightred}{rgb}{0.98,0.5,0.5}
\definecolor{lightyellow}{rgb}{1,0.90,0.40}

\usepackage{booktabs}
\usepackage{comment}

\definecolor{myyellow}{RGB}{255,255,204}
\definecolor{myred}{RGB}{255,204,204}
\definecolor{myblue}{RGB}{0,200,255}
\definecolor{mygreen}{RGB}{80,220,80}

\newcommand{\eg}{\emph{e.g.}}
\newcommand{\ie}{\emph{i.e.}}

\algblockdefx[parallel]{parallelfor}{parallelend}
[1][]{\textbf{parallel for} #1}
{\textbf{end parallel}}

\RequirePackage{mathtools}
\RequirePackage{color}
\RequirePackage{listings}

\newcommand{\ds}\displaystyle
\newcommand{\mbb}\mathbb
\newcommand{\mc}\mathcal
\newcommand{\del}\nabla

\newcommand{\beqstar}{\begin{eqnarray*}}
\newcommand{\eeqstar}{\end{eqnarray*}}

\definecolor{thegreen}{rgb}{0,.5,0}
\definecolor{idea}{rgb}{0,.6,0.1}
\definecolor{problem}{rgb}{0.7,0,0.1}
\definecolor{comment-green}{rgb}{0,.3,0}
\definecolor{theblue}{rgb}{0,0,.8}
\definecolor{light-gray}{gray}{0.98}
\definecolor{comment-color}{rgb}{0,0,.8}
\definecolor{string-color}{rgb}{0,.75,0}
\definecolor{border-blue}{rgb}{0,0,.6}

\usepackage{epsfig}

\newcommand{\ra}[1]{\renewcommand{\arraystretch}{#1}} 

\usepackage{ragged2e}

\usepackage{setspace}

\graphicspath{{./}{./images/}}
\newcolumntype{H}{>{\setbox0=\hbox\bgroup}c<{\egroup}@{}}

\definecolor{orange}{rgb}{1,0.5,0}
\definecolor{gray}{RGB}{20,20,20}
\definecolor{greencm}{RGB}{0,153,0}

\definecolor{gray}{RGB}{150,150,150}
\definecolor{theblue}{RGB}{0, 20, 159} 

\definecolor{thedarkblue}{RGB}{0,0,120} 
\usepackage{hyperref}

\definecolor{mydarkblue}{rgb}{0,0.08,0.45} 
\hypersetup{
pdftitle={},
pdfauthor={},
pdfsubject={Machine Learning},
pdfkeywords={},
pdfborder=0 0 0,
pdfpagemode=UseNone,
colorlinks=true,
linkcolor=mydarkblue,
citecolor=mydarkblue,
filecolor=mydarkblue,
urlcolor=mydarkblue,
pdfview=FitH}

\usepackage{ifpdf}
\usepackage{paralist}
\usepackage{tabularx}
\usepackage{booktabs}
\usepackage{multirow}
\usepackage{graphicx}
\usepackage{sidecap}
\usepackage{mathdots}
\usepackage{lscape}
\makeatletter
\def\vcdots{\vbox{\baselineskip4\p@ \lineskiplimit\z@
\kern3\p@\hbox{.}\hbox{.}\hbox{.}\kern3\p@}}
\makeatother

\usepackage{subfigure}
\usepackage{array}
\usepackage{url}

\newcommand{\ourMethod}{\ensuremath{\mathrm{\text{our method}}}}

\newcommand{\deepwalk}{\textrm{DeepWalk}}

\providecommand{\Y}{\ensuremath{W}} 
\providecommand{\y}{\ensuremath{w}}

\providecommand{\RR}{\mathbb{R}}

\providecommand{\Std}{\mathbb{S}}

\renewcommand{\argmax}{\operatornamewithlimits{\arg \; \max}}

\renewcommand{\phi}{\ensuremath{\Phi}}

\providecommand{\E}{\ensuremath{E}}

\providecommand{\p}{\ensuremath{p}}

\providecommand{\x}{\ensuremath{\sx}}  
\providecommand{\y}{\ensuremath{\sy}}

\providecommand{\p}{\ensuremath{p}}

\renewcommand{\O}{\ensuremath{\mathcal{O}}}
\renewcommand{\bigO}[1]{\ensuremath{\mathcal{O}(#1)}}

\providecommand{\m}{\ensuremath{M}} 
\providecommand{\n}{\ensuremath{N}}

\providecommand{\nW}{\ensuremath{M}}  
 
\providecommand{\nK}{\ensuremath{K}}  
\renewcommand{\k}{\ensuremath{K}}  

\providecommand{\nL}{\ensuremath{L}}  
\providecommand{\nR}{\ensuremath{R}}  

\providecommand{\nD}{\ensuremath{D}}

\providecommand{\N}{\ensuremath{\Gamma}}

\def\keywordname{{\bfseries Keywords}}
\def\keywords#1{\par\addvspace\medskipamount{\rightskip=0pt plus1cm
\def\and{\ifhmode\unskip\nobreak\fi\ $\cdot$
}\noindent\keywordname\enspace\ignorespaces#1\par}\vspace{5mm}}

\begin{document}

\title{A Framework for Generalizing Graph-based Representation Learning Methods}

\def\fontsz{}

\newcommand{\authorEmail}[1]{}

\author{
Nesreen K. Ahmed\\
Intel Labs\\
\authorEmail{nesreen.k.ahmed@intel.com}
\And\fontsz
Ryan A. Rossi\\
PARC\\
\authorEmail{rrossi@parc.com}
\And
Rong Zhou\\
\fontsz Google\\
\authorEmail{rongzhou@google.com}
\And
John Boaz Lee\\
\fontsz WPI\\
\authorEmail{jtlee@wpi.edu}
\AND
Xiangnan Kong\\
\fontsz WPI\\
\authorEmail{xkong@wpi.edu}
\And
\fontsz
Theodore L. Willke\\
Intel Labs \\
\authorEmail{ted.willke@intel.com}
\And
Hoda Eldardiry\\
\fontsz
PARC\\
\authorEmail{heldardiry@parc.com}
}

\maketitle

\begin{abstract}
\vspace{-1mm}
Random walks are at the heart of many existing deep learning algorithms for graph data.
However, such algorithms have many limitations that arise from the use of random walks, \eg, the features resulting from these methods are unable to transfer to new nodes and graphs as they are tied to node identity.
In this work, we introduce the notion of \emph{attributed random walks} which serves as a basis for generalizing existing methods
such as DeepWalk, node2vec, and many others that leverage random walks.
Our proposed framework enables these methods to be more widely applicable for both transductive and inductive learning as well as for use on graphs with attributes (if available). This is achieved by learning functions that generalize to new nodes and graphs.
We show that our proposed framework is effective with an average AUC improvement of $16.1\%$ while requiring on average 853 times less space than existing methods on a variety of graphs from several domains.

\end{abstract}

\section{Introduction}
\label{sec:intro}
Graphs (networks) are ubiquitous~\cite{borgatti1992notions,akoglu2015graph} and allow us to model entities (nodes) and the dependencies (edges) between them.
Graph data is often observed directly in the natural world (\eg, biological or social networks)~\cite{viswanath2009evolution} or constructed from non-relational data by deriving a metric space between entities and retaining only the most significant edges~\cite{zhu2003semi,henaff2015deep,rossi2012transforming}.
Learning a useful feature representation from graph data lies at the heart and success of many machine learning tasks such as
node classification~\cite{neville2000iterative,mcdowell2009cautious},
anomaly detection~\cite{akoglu2015graph},
link prediction~\cite{al2011survey},
dynamic network analysis~\cite{nicosia2013graph}, 
community detection~\cite{radicchi2004defining,ng2002spectral}, 
role discovery~\cite{rossi2015-tkde}, 
visualization and sensemaking~\cite{gvis-icwsm15,pienta2015scalable}, 
graph classification~\cite{kearnes2016molecular,skipgraph}, and
network alignment~\cite{koyuturk2006pairwise}.

Many existing techniques use \emph{random walks} as a basis for learning features or estimating the parameters of a graph model for a downstream prediction task.
Examples include recent node embedding methods such as DeepWalk~\cite{deepwalk}, node2vec~\cite{node2vec}, as well as graph-based deep learning algorithms.
However, the simple random walk used by these methods is fundamentally tied to the \emph{identity} of the node.
This has three main disadvantages.
First, these approaches are inherently transductive and do not generalize to unseen nodes and other graphs.
Furthermore, they are unable to be used for graph-based transfer learning tasks such as across-network classification~\cite{kuwadekar2011relational,introSRL07}, 
graph similarity~\cite{goldsmith1990assessing,zager2008graph}, and
matching~\cite{prvzulj2007biological,park2010fast}.
Second, they are not space-efficient as a feature vector is learned for each node which is impractical for large graphs.
Third, most of these approaches lack support for \emph{attributed graphs}. 
This includes graphs with intrinsic (node) attributes such as age or gender as well as structural features such as higher-order subgraph counts, \eg, number of 4-cliques each node participates.

To make these methods more generally applicable, 
we introduce a reinterpretation of the notion of random walk that is not tied to node identity and is instead based on a function $\phi : \vx \rightarrow \y$ that maps a node 
attribute vector to a type.
Using this reinterpretation as a basis we propose a framework that naturally generalizes many existing methods.
This framework provides a number of important advantages.
First, the learned features generalize to new nodes and across graphs and therefore can be used for transfer learning tasks such as across-network link prediction and classification.
Our proposed approach supports both transductive and inductive learning.
Second, the generalized approach is inherently space-efficient since embeddings are learned for types (as opposed to nodes) and therefore requires significantly less space than existing methods.
Third, the generalized approach supports learning from attributed graphs.
However, we stress that the approach does not require graphs with input attributes since these can be derived from the graph structure.
Furthermore, our approach is shown to be effective with an average improvement of $16.1\%$ in AUC while requiring on average $853$x less space than existing methods on a variety of graphs.

\smallskip\noindent
\textbf{Contributions}:
This paper proposes a framework for generalizing many existing algorithms making them more broadly applicable for attributed graphs and inductive learning.
It has the following key properties:

{
\begin{compactitem}
\setlength{\parskip}{1mm}
\item \textbf{Space-efficient}: It requires on average $853$x less space than a number of existing methods.

\item \textbf{Accurate}: It is accurate with an average improvement of $16.1\%$ across a variety of graphs from several domains.

\item \textbf{Inductive}: 
It is an inductive learning approach that is able to learn embeddings for 
new nodes and graphs.

\item \textbf{Attributed}: 
It naturally supports graphs with attributes (if available) and serves as a foundation for generalizing existing methods for use on attributed graphs.

\vspace{1.5mm}
\end{compactitem}
}

\section{Framework} 
\label{sec:framework}
In this section, we formally introduce the attributed random walk framework which can serve as a basis for generalizing many existing deep graph models and embedding methods.
The framework consists of two general components:
\begin{enumerate}
\item[$\mathbf{C1}$.] (\textbf{Function Mapping Nodes to Types}):
A function $\phi$ that maps nodes to types based on a $\n \times \k$ matrix $\mX$ of attributes.
The function $\phi$ can be defined or learned automatically.
Note $\mX$ may be given as input or computed based on the structure of the graph.
For more details, see Section~\ref{sec:framework-function-mapping-nodes-to-types}. 

\item[$\mathbf{C2}$.] (\textbf{Attributed Random Walks}): 
Generate \emph{attributed random walks} based on the function $\phi$ mapping nodes to types (Section~\ref{sec:framework-attributed-random-walks}).
Informally, an attributed walk is simply the node types that occur during a walk.
\end{enumerate}
A summary of the notation is provided in Table~\ref{table:notation}.
To avoid abuse of notation, this paper presents the framework for nodes.
However, the approach is applicable to both nodes and edges.

\begin{table}[b!]
\vspace{-3mm}
\centering
\caption{Summary of notation.
}
\vspace{1mm}
\setlength\extrarowheight{0pt}
\centering
\centering 
\small
\footnotesize
\fontsize{7.5}{8.2}\selectfont
\fontsize{8}{8.5}\selectfont
\setlength{\tabcolsep}{6pt} 
\label{table:notation}
\def\arraystretch{1.18}
\begin{tabularx}{1.0\linewidth} 
{@{}r
X@{}} 
\toprule

$G$ & (un)directed (attributed) graph \\

$\n$ & number of nodes $|V|$ \\

$\nW$ & number of unique types 
\\ 

$\nK$ & number of attributes used for deriving types 
\\

$\nD$ & number of dimensions to use for embedding\\

$\nR$ & number of walks for each node \\ 
$\nL$ & length of a random walk \\

$\N_{i}$ & neighbors (adjacent nodes) of $v_i$ \\

$\mathcal{S}$ & set of (attributed) random walks \\

$\phi(\cdot)$ & 
a function $\phi : \vx \rightarrow \y$ that maps an attribute vector $\vx \in \RR^{\nK}$ to a corresponding type
\\

$\Y$ & 
set of types where $\y \in \Y$ is a type assigned to one or more nodes
\\

$\mX$  & 
an $\n \times \k$ attribute matrix where the rows represent nodes and the columns represent attributes 
\\

$\vx_i$ & a $\nK$-dimensional attribute vector for node $v_i$ \\

$\bar{\vx}_j$ & a $\n$-dimensional vector for attribute $j$ (column of $\mX$) \\

$\alpha$ & transformation hyperparameter 
\\

$\mZ$ & 
an $\m \times \nD$ matrix of type embeddings
\\

$\vz_k$ & a $\nD$-dimensional embedding for type $\y_k$
\\

\bottomrule
\end{tabularx}
\vspace{-2mm}
\end{table}

\subsection{Function Mapping Nodes to Types}
\label{sec:framework-function-mapping-nodes-to-types}
Given an (un)directed graph $G=(V,E)$ and an $\n \times \k$ matrix $\mX$ of attributes (if available), 
the first component of the framework maps the $\n$ nodes to a set $\Y = \{ \y_{1},...,\y_{\m} \}$ of $\m$ types 
where $1 \leq \m \leq \n$ and $\m$ is often much smaller than $\n$, \ie, $\m \ll \n$.
More formally, 
\begin{gather}
\phi \; : \; \vx \; \rightarrow \; \y
\end{gather}
\noindent
where $\phi$ is a function mapping nodes to \emph{types} based on the $\n \times \k$ attribute matrix $\mX$.
The function $\phi$ can be learned automatically or defined manually by the user.
The framework is general and flexible for use with any arbitrary function $\phi$ that maps nodes to types based on an $\n \times \k$ attribute matrix $\mX$.
For graphs that are not attributed, we simply derive $\mX$ by extracting a set of graph features based on the structure of the graph (\eg, $\mX$ may represent the graphlet features shown in Figure~\ref{fig:graphlet-attributes}).\footnote{Graphlets (and orbits) can be derived using exact algorithms~\cite{ahmed2017kais} or accurately estimated using fast parallel approximation methods with provable error bounds~\cite{ahmed16bigdata,rossi17graphlet-est}.}
The set of types $\Y$ is defined as:
\begin{gather}\label{eq:set-of-words-node-functions}
\Y = \, \phi(\vx_1) \, \cup \, \phi(\vx_2) \, \cup \, \cdots \, \cup \, \phi(\vx_{\n})
\end{gather}
\noindent 
where
$\phi$ is a function that maps the set of $\n=|V|$ nodes to $\m=|\Y|$ types such that $1\leq \m \leq \n$.
In node2vec and other embedding and deep graph algorithms, notice that each node is mapped to a unique identifier that identifies the specific node which is then used in the random walk.
However, in the attributed random walk framework two or more nodes may map to the same type.

\begin{figure*}[t!]
\centering
\includegraphics[width=0.75\linewidth]{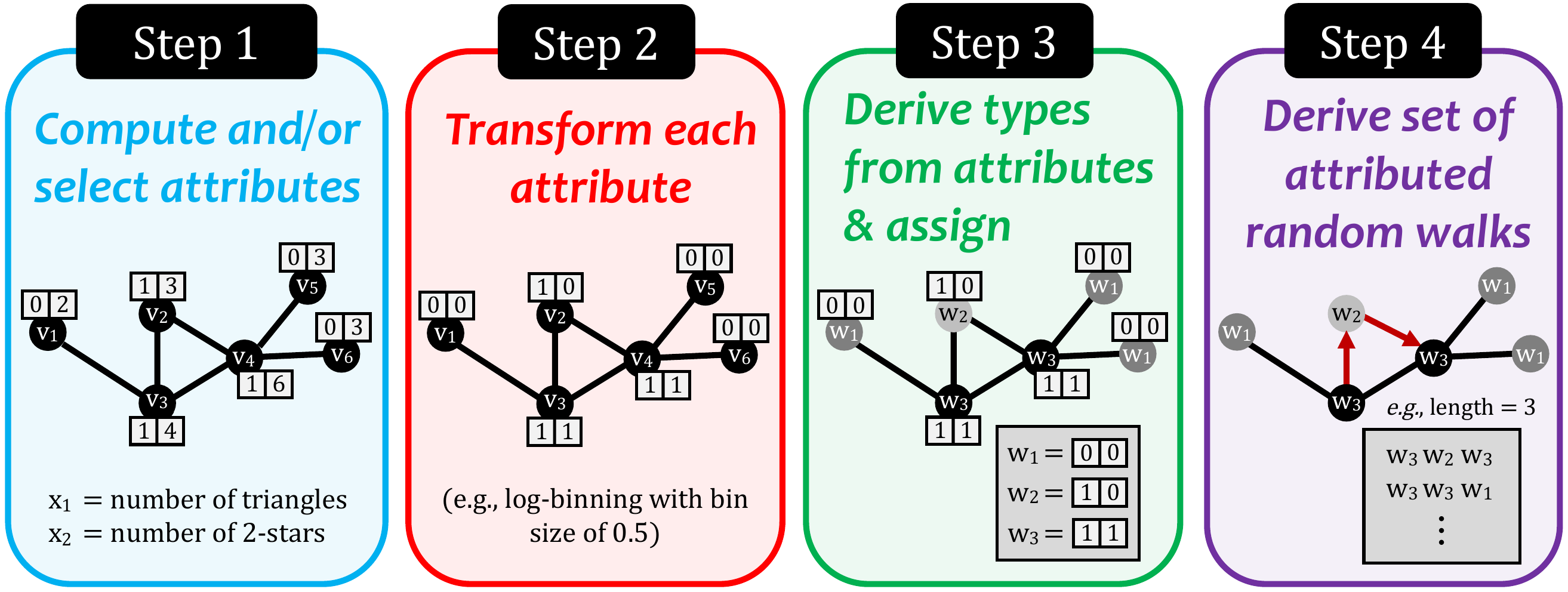}
\vspace{-1mm}
\caption{
A simple illustrative toy example.
This toy example shows only one potential instantiation of the general framework.
In particular, Step 1 computes the number of triangles and 2-stars that each node participates whereas 
Step 2 transforms each individual attribute using logarithmic binning.
Next, Step 3 derives types using 
a simple function $\phi$ representing a concatenation of each node's attribute values resulting in 
$\m=3$ types 
$\{\y_1, \y_2, \y_3\}$ 
among the $\n=6$ nodes.
Finally, Step 4 derives a set of attributed random walks 
(type sequences) 
which are then used to learn embeddings.
See Section~\ref{sec:illustrative-example} for further discussion.
}
\label{fig:overview}
\end{figure*}

There are two general classes of functions for mapping nodes to types.
The first class of functions are simple functions taking the form:
\begin{gather}\label{eq:simple-function}
\phi(\vx) = x_1 \circ x_2 \circ \cdots \circ x_{\k}
\end{gather} 
\noindent
where $\vx$ is an attribute vector:
\begin{gather}\label{eq:attr-vector}
\vx = \big[\; x_1\;\, x_2\;\, \cdots \;\, x_{\k} \; \big]
\end{gather}
\noindent
and $\circ$ is a 
binary operator such as concatenation, sum, among others.
The second class of functions are learned by solving an objective function.
This includes functions based on a low-rank factorization of the $\n \times \k$ matrix $\mX$ having the form $\mX \approx f\langle\mU\mV^T\rangle$ with factor matrices $\mU \in \RR^{\n \times F}$ and $\mV \in \RR^{\k \times F}$ where $F>0$ is the rank and $f$ is a linear or non-linear function. 
More formally,
\begin{align}\label{eq:low-rank}
\arg\min_{\mU,\mV \in \mathcal{C}} \; 
\Big[
\mathcal{D} \big(\mX, f\langle\mU\mV^T\rangle \big) + \mathcal{R}(\mU, \mV) 
\Big]
\end{align}
\noindent
where $\mathcal{D}$ is the loss, $\mathcal{C}$ is constraints (\eg, non-negativity constraints $\mU \geq 0, \mV \geq 0$), and $\mathcal{R}(\mU, \mV)$ is a regularization penalty.
Let $V_k$ denote the set of nodes mapped to type $w_k \in W$.
We partition $\mU \in \RR^{\n \times F}$ into $\m$ disjoint sets of nodes (for each of the $\m$ types) $V_1, \ldots, V_{\m}$ such that $V = V_1 \cup \ldots \cup V_{\m}$ by solving the k-means objective:
\begin{align}\label{eq:k-means-objective-func}
\min_{\{V_j\}^{\m}_{j=1}} \sum_{j=1}^{\m} \sum_{\vu_i \in V_j} \| \vu_i - \vc_j \|^2,
\text{where } \vc_j = \frac{\sum_{\vu_i \in V_j} \vu_i}{|V_j|}
\end{align}
Alternatively, the nodes can be mapped to types directly by setting $F=\m$ and using $\mU$ as follows:
\begin{align}\label{eq:direct-types-from-U}
\phi(\vx_i) = \argmax_{\y_k \in \Y} \; U_{ik}, \quad \forall\, i=1, \ldots, \n
\end{align}
For other possibilities, we refer the readers to a recent survey by Xu~\etal~\cite{xu2005survey}.
Notice that the approaches in Eq.~\eqref{eq:low-rank}-\eqref{eq:direct-types-from-U} can be used for inductive learning as follows:
Given a new graph $G^{\prime}=(V^{\prime}, E^{\prime})$ (or node $v_{\n+1}$), 
we compute the attribute matrix $\mX^{\prime}$ (or attribute vector $\vx_{\n+1}$) using the same $\k$ attributes 
and then 
use $\mX^{\prime}$ and 
$\mV$ from Eq.~\eqref{eq:low-rank} to estimate $\mU^{\prime}$ directly.
More formally, 
given $\mX^{\prime}$ and $\mV$, we find $\mU^{\prime}$ by solving:
\begin{align}\label{eq:low-rank-estimate-Uprime-given-Xprime-and-V}
\argmin_{\mU^{\prime} \in \mathcal{C}} \; 
\Big[
\mathcal{D} \big(\mX^{\prime}, f\langle\mU^{\prime}\mV^{T}\rangle \big) + \mathcal{R}(\mU^{\prime})
\Big]
\end{align}
\noindent
Observe that $\mV$ gives the mapping from the latent feature space to the input attribute space and represents how each of the $\k$ attributes map to the latent features and therefore $\mV$ is not tied to particular nodes but more generally to the (structural) attributes. 
Many of these methods avoid 
the issue of selecting an appropriate subset of attributes to use apriori and 
are more robust
to the selection of different sets of attributes compared to simple functions such as $\phi(\vx) = x_1 \circ x_2 \circ \cdots \circ x_{\k}$.
As an aside, it is straightforward to formulate a function learning problem to automate the choice of function $\phi$.

The framework naturally supports both attributed and non-attributed graphs as the attribute matrix $\mX$ can be derived by extracting a set of graph features based on the structure of the graph (\eg, $\mX$ may represent the graphlet features shown in Figure~\ref{fig:graphlet-attributes}) or learned automatically using an inductive feature learning approach such as DeepGL~\cite{deepGL}. 

\subsection{Attributed Random Walks}
\label{sec:framework-attributed-random-walks}
Most existing deep learning models and embedding algorithms for graphs use simple random walks based on node identity~\cite{deepwalk,node2vec}.
The features learned from these methods are fundamentally tied to the identity of a node and therefore are unable to generalize to new nodes as well as for graph-based transfer learning tasks or across-network learning, among other limitations.
In this section, we introduce the notion of attributed random walks and demonstrate 
how they serve as a basis
for generalizing many existing deep graph models and embedding algorithms.

Previous work uses random walks based on the following traditional definition of a walk:
\begin{mydef}[Walk] \label{def:simple-walk}
A walk $S$ of length $\nL$ is defined as a sequence of indices $i_{1}, i_{2}, \ldots, i_{\nL+1}$ such that $(v_{i_{t}}, v_{i_{t+1}}) \in E$ for all $1 \leq t \leq \nL$.
\end{mydef}
Note that nodes in a walk $S$ are not necessarily distinct.
We replace the traditional notion of walk (Definition~\ref{def:simple-walk}) with the more appropriate and useful notion of \emph{attributed walk} defined as:
\begin{mydef}[Attributed walk]\label{def:attr-walk}
Let $\vx_i$ be a $\k$-dimensional vector for node $v_i$.
An \emph{attributed walk} $S$ of length $\nL$ is defined as a sequence of adjacent node types 
\begin{gather} \label{eq:attr-random-walk}
\phi(\vx_{i_{1}}), \phi(\vx_{i_{2}}),\ldots, \phi(\vx_{i_{\nL+1}})
\end{gather}
associated with a sequence of indices $i_{1}, i_{2}, \ldots, i_{\nL+1}$ such that $(v_{i_{t}}, v_{i_{t+1}}) \in E$ for all $1 \leq t \leq \nL$ and
$\phi : \vx \rightarrow \y$ is a function that maps the input vector $\vx$ of a node to a corresponding type $\phi(\vx)$.
\end{mydef}
The type sequence $\phi(\vx_{i_{1}}), \phi(\vx_{i_{2}}),\ldots, \phi(\vx_{i_{\nL+1}})$ is simply the node types that occur during a walk.
\noindent
The framework allows both uniform and non-uniform attributed random walks.
It is also straightforward to bias the attributed random walk in an arbitrary fashion.
The notion of attributed walk can be extended in various ways (\eg, for edge types).

Given an arbitrary embedding method or deep graph model $\mathcal{A}$ that uses simple random walks based on node identity (\eg, node2vec, DeepWalk, among others), 
we generalize $\mathcal{A}$ by replacing the traditional notion of random walk with the proposed notion of \emph{attributed random walks}.
We will show that existing methods are actually a special case of the proposed framework when $\phi(\vx_i)$ uniquely identifies node $v_i$.
Suppose the graph $G$ has $\n$ nodes mapped to $\m$ types using an arbitrary function $\phi$ (Section~\ref{sec:framework-function-mapping-nodes-to-types}).
If we select $\phi$ such that $\m \rightarrow \n$ then we recover the original method $\mathcal{A}$ as a special case of the framework.
More intuitively, each node $v_i \in V$ must be assigned to a unique type $\phi(\vx_i)=\y_i$ that uniquely identifies it from any other node $v_j \in V$ since $|\Y|=|V|$ where $\Y = \phi(\vx_1) \cup \cdots \cup \phi(\vx_{\n})$.
Observe that $\m \rightarrow \n$ is actually an extreme case of the framework and corresponds exactly to the original method $\mathcal{A}$.

There are three key advantages that arise when existing methods are generalized by the proposed framework.
First, the generalized method naturally supports attributed graphs.
Second, the features 
learned generalize to new nodes and across graphs and therefore are naturally inductive and able to be used for transfer learning tasks.
Finally, the learned embeddings are significantly more space-efficient with a space complexity of only $\O(\m\nD)$ compared to previous methods that require $\O(\n\nD)$ space where $\m \ll \n$ ($\m$ is much smaller than $\n$) and $\nD$ is the embedding dimensionality.

\subsection{Illustrative Example}
\label{sec:illustrative-example}
We emphasize that the proposed framework 
is general and flexible; and the example discussed in this section represents only one such 
simple instantiation of the framework.
An overview of the key steps involved in this specific instantiation 
are shown in Figure~\ref{fig:overview} and summarized succinctly below as follows:

\begin{enumerate}

\item[$\mathbf{Step\; 1.}$]
\textbf{Compute and/or Select Attributes}: 
Given an (un)directed graph $G$, 
the first step is to compute a set of attributes using the graph structure (\eg, graphlets).

\item[$\mathbf{Step\; 2.}$]
\textbf{Transform Attributes}: 
Next, we transform each attribute vector.
The goal is to map the values of each individual 
attribute vector $\bar{\vx}_j$ to a smaller set of values (via quantization or discretization algorithm).
In Figure~\ref{fig:overview}, each attribute $\bar{\vx}_j$ is transformed using logarithmic binning.\footnote{Logarithmic binning assigns the first $\alpha\n$ nodes with smallest attribute value to $0$ (where $0<\alpha<1$), then assigns the $\alpha$ fraction of remaining unassigned nodes with smallest value to $1$, and so on.}
For convenience, each initial $\bar{\vx}_j$ is replaced by the transformed attribute.

\item[$\mathbf{Step\; 3.}$]
\textbf{Derive Types from Attributes \& Assign}: 
Now, the set of 
\emph{types} are derived as 
$\Y = \phi(\vx_1) \cup \phi(\vx_2) \cup \cdots \cup \phi(\vx_{\n})$
where $|\Y| \leq \n$ and 
$\phi : \vx \rightarrow \y$ such that 
$\phi(\vx) = x_1 \circ x_2 \circ \cdots \circ x_{\k}$ and $\circ$ is a binary operator such as concatenation.
We also assign each node to its corresponding type. See Figure~\ref{fig:overview} for intuition.

\item[$\mathbf{Step\; 4.}$]
\textbf{Attributed Random Walks}: 
Finally, we generate a set of attributed random walks 
using the node types (as opposed to the unique node identifiers used traditionally).
The attribute random walks represent sequences of node types 
which are then used to learn embeddings.
For instance, the set of attributed random walks are used in Section~\ref{sec:node2vec-generalization} to learn the $\m \times \nD$ type embedding matrix $\mZ$.
\end{enumerate}
\noindent
From the perspective of the general framework, 
Steps 1-3 in Figure~\ref{fig:overview} correspond to the first main component that maps nodes to types via the function $\phi$ while Step 4 corresponds to the last component that generates \emph{attributed random walks} representing sequences of node types.

{\algrenewcommand{\alglinenumber}[1]{\fontsize{6.5}{7}\selectfont#1 }
\newcommand{\multiline}[1]{\State \parbox[t]{\dimexpr\linewidth-\algorithmicindent}{#1\strut}}
\begin{figure}[t!]
\vspace{-2mm}
\centering
\begin{algorithm}[H]
\caption{\,\small Generalized Node2Vec
\vspace{0.2mm}
}
\label{alg:generalized-node2vec}{
\begin{spacing}{1.3}
\fontsize{8}{9}\selectfont
\begin{algorithmic}[1]
\Procedure{GeneralizedNode2Vec}{
$G = (V,\E)$ and possibly attributes $\mX$,
embedding dimensions $\nD$, 
walks per node $\nR$, walk length $\nL$, 
context (window) size $\omega$, 
return $p$, and in-out $q$, 
function $\phi$
}
\smallskip

\State Initialize set of \emph{attributed walks} $\mathcal{S}$ to $\emptyset$ \label{algline:node2vec-gen-init-attr-walks}

\State Extract (graphlet) features if needed and append to $\mX$ \label{algline:node2vec-gen-extract-features}
\State Transform each attribute in $\mX$ (\eg, using logarithmic binning) \label{algline:nod2vec-gen-transform-features}

\State Precompute transition probabilities $\pi$ using $G$, $p$, and $q$
\label{algline:node2vec-gen-preprocess-mod-weights}

\State $G^{\prime} = (V,E,\pi)$ 

\parfor[$j = 1,2,...,\nR$] \label{algline:node2vec-gen-for-each-random-walk} \Comment{walks per node} 
	\State Set $\Pi$ to be a random permutation of the nodes $V$
	
	\For{{\bf each} $v \in \Pi$ in order} 
	
		\State $S = \textsc{AttributedWalk}(G^{\prime}, \mX, v, \Phi, \nL)$ 
		\label{algline:node2vec-gen-obtain-attr-walk}
		\State Add the \emph{attributed walk} $S$ to $\mathcal{S}$ \label{algline:node2vec-gen-add-attr-walk-to-set}
	\EndFor
\endpar
\State $\mZ = \textsc{StochasticGradientDescent}(\omega, \nD, \mathcal{S})$ \label{algline:node2vec-gen-SGD-with-attr-walks} 
\Comment{$\mathbf{parallel}$}
\State \textbf{return} the learned \emph{type} embeddings $\mZ$ \label{algline:node2vec-gen-return-learned-representation-matrix} 
\EndProcedure
\end{algorithmic}
\end{spacing}}
\end{algorithm}
\vspace{-4mm}
\end{figure}}

{\algrenewcommand{\alglinenumber}[1]{\fontsize{6.5}{7}\selectfont#1 }
\newcommand{\multiline}[1]{\State \parbox[t]{\dimexpr\linewidth-\algorithmicindent}{#1\strut}}
\begin{figure}[t!]
\vspace{-2mm}
\begin{algorithm}[H]
\caption{\,\small Attributed Random Walk
\vspace{0.2mm}
}
\label{alg:gen-node2vec-attr-walk}{
\begin{spacing}{1.3}
\fontsize{8}{9}\selectfont
\begin{algorithmic}[1]
\Procedure{AttributedWalk}{$G^{\prime}$, $\mX$, start node $s$, function $\Phi$,
$\nL$}
\State Initialize attributed walk $S$ to $\big[ \Phi(\vx_s) \big]$ \label{algline:node2vec-gen-attr-walk-init-walk-and-add-start-node-function}
\State Set $i = s$ \Comment{current node} \label{algline:node2vec-gen-attr-walk-set-curr-node-to-start-node}
\For{$\ell = 1$ {\bf to} $\nL$} \label{algline:node2vec-gen-attr-walk-for}
	\State $\N_i = $ Set of the neighbors for node $i$ \label{algline:node2vec-gen-attr-walk-get-neighbors}
	\State $j = \textsc{AliasSample}(\N_i, \pi)$ \Comment{select node $j \in \N_i$} \label{algline:node2vec-gen-attr-walk-alias-sample}
	\State Append $\Phi(\vx_j)$ to $S$ \label{algline:node2vec-gen-attr-walk-add-node-function-to-list}
	\State Set $i$ to be the current node $j$ 
	\label{algline:node2vec-gen-attr-walk-curr-node-i}
\EndFor \label{algline:node2vec-gen-attr-walk-for-end}
\State \textbf{return} attributed walk $S$ of length $\nL$ rooted at node $s$ \label{algline:node2vec-gen-attr-walk-return-attr-walk}
\EndProcedure
\end{algorithmic}
\end{spacing}}
\end{algorithm}
\vspace{-4mm}
\end{figure}
}

\section{Node2Vec Generalization}
\label{sec:node2vec-generalization}
We use the proposed framework in Section~\ref{sec:framework} to generalize node2vec~\cite{node2vec}.\footnote{However, the framework could have been used to generalize any node embedding or deep graph model that leverages traditional random walks.}
This gives rise to a new generalized node2vec algorithm with the following advantages: 
(a) naturally supports attributed graphs,
(b) learns features that generalize for new nodes as well as for extraction on another arbitrary graph and thus able to be used for transfer learning tasks, and 
(c) space-efficient by learning an embedding for each type as opposed to each node.
The generalized node2vec algorithm is shown in Alg.~\ref{alg:generalized-node2vec}.
In particular, we replace the notion of random walk in node2vec with the more appropriate notion of \emph{attributed random walk} (Line~\ref{algline:node2vec-gen-obtain-attr-walk}-\ref{algline:node2vec-gen-add-attr-walk-to-set}).
Nodes are mapped to types using an arbitrary function $\Phi(\vx)$ which can be defined or learned; see Section~\ref{sec:framework-function-mapping-nodes-to-types} for further details.
As an aside, Alg.~\ref{alg:generalized-node2vec} assumes the function $\Phi$ is defined or learned apriori.
Note that if no attribute matrix $\mX$ is given as input then we derive features based on the graph structure
(Line~\ref{algline:node2vec-gen-extract-features}).
However, even if (intrinsic/self- or structural) attributes are given as input we may also choose to derive additional structural features such as graphlets~\cite{pgd} and append these to $\mX$.

The attributed random walk routine for the generalized node2vec algorithm is shown in Alg.~\ref{alg:gen-node2vec-attr-walk}.
Given a start node $s$, we first obtain the \emph{type} of the start node $s$ given by the function $\Phi(\vx_s)$ 
and initialize an attributed walk $S$ (stored as a list\footnote{A list or other efficient data structure may be used for storing a sequence of items (nodes/edges) with potential duplicates}) to be $S=\big[ \Phi(\vx_s) \big]$ (Line~\ref{algline:node2vec-gen-attr-walk-init-walk-and-add-start-node-function}).
Next, Line~\ref{algline:node2vec-gen-attr-walk-set-curr-node-to-start-node} sets the current node $i$ to be the start node $s$.
The attributed walk $S$ of length $\nL$ rooted at the start node $s$ is obtained in Lines~\ref{algline:node2vec-gen-attr-walk-for}-\ref{algline:node2vec-gen-attr-walk-for-end}.
In particular, Line~\ref{algline:node2vec-gen-attr-walk-get-neighbors} gets the neighbors of the current node $i$ which is used in Line~\ref{algline:node2vec-gen-attr-walk-alias-sample} to sample a new node $j \in \N_i$.
The type $\Phi(\vx_j)$ of node $j$ is appended to $S$ (\eg, if $S$ is stored using an ordered list then we append $\Phi(\vx_j)$ to the back of $S$ to preserve order).
Afterwards, Line~\ref{algline:node2vec-gen-attr-walk-curr-node-i} sets the current node $i$ to be $j$ and Lines~\ref{algline:node2vec-gen-attr-walk-for}-\ref{algline:node2vec-gen-attr-walk-for-end} are repeated for $\nL$ steps. 
Finally, an attributed random walk $S$ of length $\nL$ starting at node $s$ is returned in Line~\ref{algline:node2vec-gen-attr-walk-return-attr-walk}.
One can also leverage Alg.~\ref{alg:gen-node2vec-attr-walk} (or an adapted variant of it) for generalizing other deep graph models and node embedding methods.
It is straightforward to see that Alg.~\ref{alg:generalized-node2vec} is a generalization of node2vec since if we allow $\m \rightarrow \n$ then we recover the original node2vec algorithm as a special case of the framework.
Furthermore, $\deepwalk$~\cite{deepwalk} is also a special case where 
$\m \rightarrow \n$, $p=1$, and $q=1$.

\smallskip
\noindent
\textbf{Implementation details:}
In our implementation, we evaluate $\Phi(\vx_i) \,=\, x_1 \circ \, x_2 \, \circ \, \cdots \, \circ \, x_{\nK}$ only once for each node $i \in V$
and store each $\Phi(\vx_i)$ in a hash table giving $o(1)$ time lookup.
This allows us to derive $\Phi(\vx_s)$ and $\Phi(\vx_j)$ in Line~\ref{algline:node2vec-gen-attr-walk-init-walk-and-add-start-node-function} and~\ref{algline:node2vec-gen-attr-walk-add-node-function-to-list} of Alg.~\ref{alg:gen-node2vec-attr-walk} in $o(1)$ constant time.
To construct the hash table it takes only $\O(\n \nK)$ time to obtain $\Phi(\vx_i), \forall i=1,...,\n$ given $\mX$ and 
only $\O(\n)$ space to store them efficiently using a hash table.

\section{Experiments} 
\label{sec:exp}
In this section, we investigate the effectiveness of the proposed framework using a variety of graphs from several domains.
In these experiments, we use the node2vec generalization given in Section~\ref{sec:node2vec-generalization}. 

\begin{figure}[h!]
\centering
\includegraphics[width=0.75\linewidth]{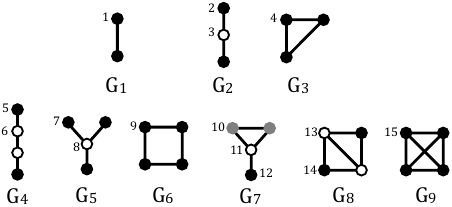}
\vspace{-1mm}
\caption{
Summary of the 9 node graphlets and 15 orbits (graphlet automorphisms) with 2-4 nodes.
}
\label{fig:graphlet-attributes}
\end{figure}

\subsection{Experimental Setup} 
\label{sec:exp-settings}
Unless otherwise mentioned, all experiments use logarithmic binning (defined in Section~\ref{sec:illustrative-example}) with $\alpha=0.5$.
In these experiments, we use a simple function $\phi(\vx)$ that represents a concatenation of the attribute values in the node attribute vector $\vx$.
In particular, $\phi(\vx)$ is defined as $\phi(\vx) = x_1 \, \circ \, \cdots \, \circ \, x_\k$ where $\circ$ is the concatenation operator.
For these experiments, we searched over 10 subsets of the $9$ graphlet attributes shown in Figure~\ref{fig:graphlet-attributes}.\footnote{Note we also investigated the 15 graphlet orbits (as opposed to the 9 graphlet attributes) 
and found similar results.}
However, the approach trivially handles other categories of attributes including structural attributes, intrinsic/self-attributes (such as age or other non-relational attributes), and relational features derived using the graph $G$ along with existing structural or self-attributes.
We evaluate the \emph{generalized node2vec} approach presented in Section~\ref{sec:node2vec-generalization} that leverages the attributed random walk framework (Section~\ref{sec:framework}) against a number of popular methods including:
node2vec\footnote{$\mathtt{https}$://$\mathtt{github.com/aditya}$-$\mathtt{grover/node2vec}$}, 
DeepWalk~\cite{deepwalk}, and 
LINE~\cite{line}.
For our approach and node2vec, we use the same hyperparameters ($\nD=128$, $\nR=10$, $\nL=80$) and grid search over $p,q\in \{0.25, 0.50, 1, 2, 4\}$ as mentioned in~\cite{node2vec}.
We use logistic regression (LR) with an L2 penalty.
The model is selected using 10-fold cross-validation on $10\%$ of the labeled data.
Experiments are repeated for 10 random seed initializations.
All results are statistically significant with p-value $< 0.01$.
Unless otherwise mentioned, we use AUC to evaluate the models.
Data has been made available at NetworkRepository~\cite{nr}.\footnote{$\mathtt{http}$://$\mathtt{networkrepository.com/}$\!}

\begin{figure}[h!]
\centering

\vspace{-4mm}
\subfigure[27 types ($\phi$ only)]{
\includegraphics[width=0.31\linewidth]{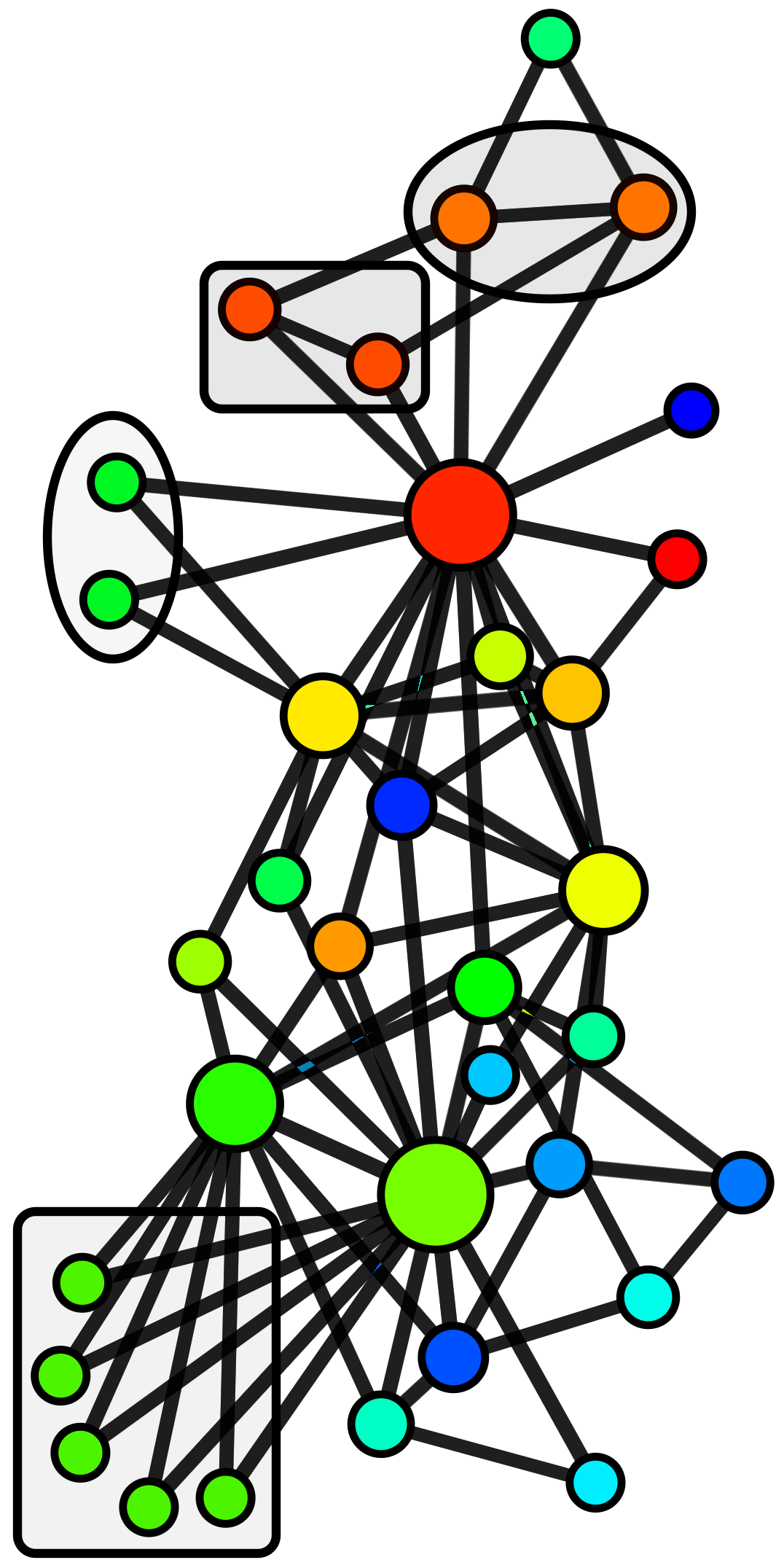}
\label{fig:attr-tri-2stars-no-trans}}
\hspace{4mm}
\subfigure[7 types (LB, $\phi$)]
{\includegraphics[width=0.30\linewidth]{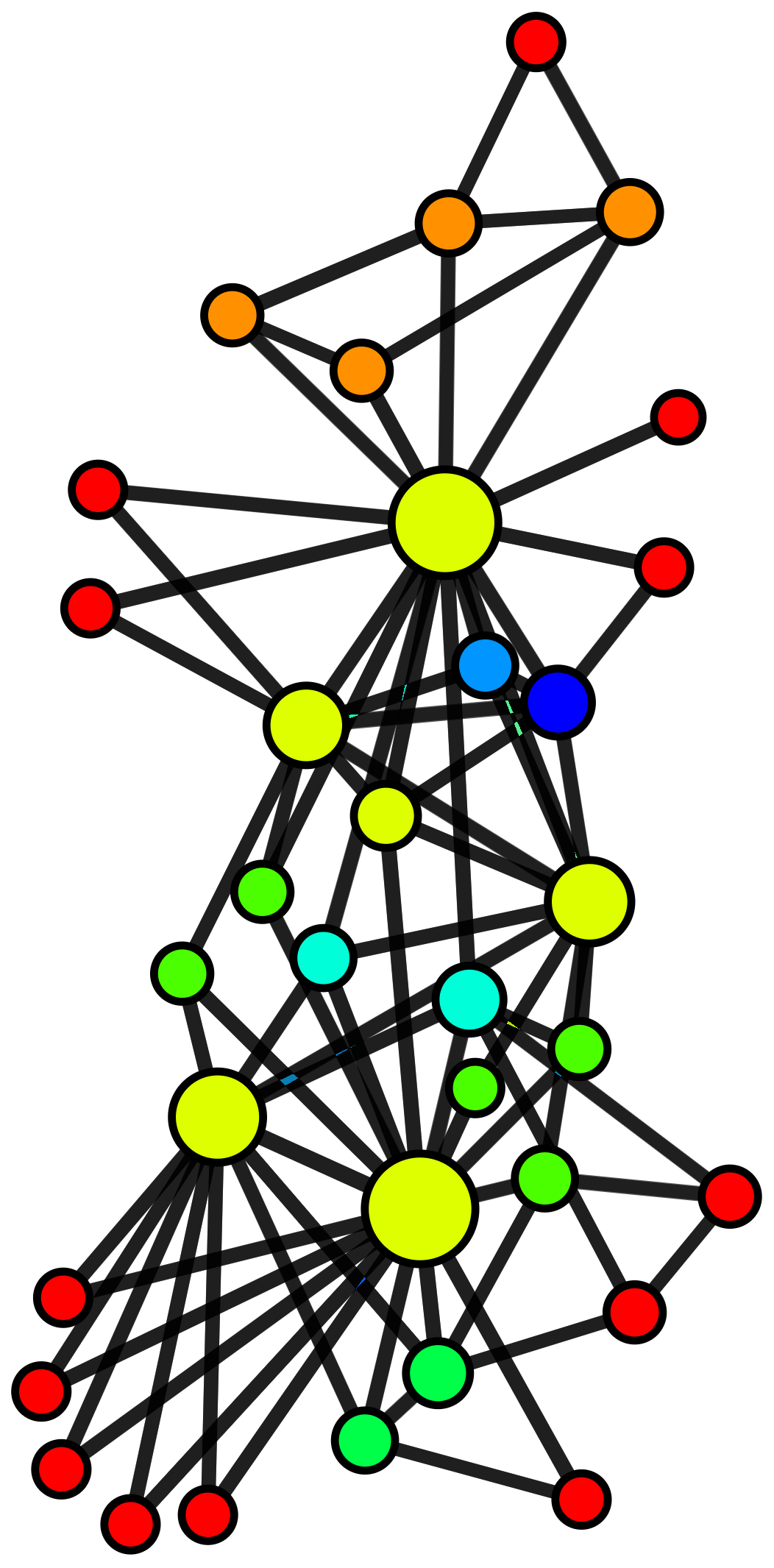}
\label{fig:attr-tri-2stars-logbinning}}

\vspace{-1mm}
\caption{
Using only the node types are useful for revealing interesting insights into the graph data including nodes that are structurally similar.
Node types alone reveal nodes that are structurally similar (karate network).
Node color encodes type.
Only the number of triangles and wedges are used by $\phi$.
In (a) nodes with identical types are grouped; 
(b) uses log binning (LB) with $\alpha=0.5$. 
See text for further discussion.
}
\label{fig:attrWords-noTrans-vs-LogBinning}
\end{figure}

\subsection{Node Mapping Experiments}
\label{sec:exp-use-case}
This section investigates the intermediate representation defined by an arbitrary function $\phi$ that maps nodes to a set of types. 
Recall that for simplicity 
$\phi$ is defined as a concatenation of the attribute values for a given node and the attributes correspond to the graphlet counts in Figure~\ref{fig:graphlet-attributes}.
Strikingly, we observe in Figure~\ref{fig:attrWords-noTrans-vs-LogBinning} that even these simple mappings alone are useful and effective for understanding the important
structural and behavioral characteristics of nodes, that is before even generating attributed random walks and learning an embedding based on them.
In particular, nodes assigned to the same type 
in Figure~\ref{fig:attr-tri-2stars-no-trans} obey a strict notion of node equivalence on a feature representation: 
\begin{align}\label{eq:node-equiv-feature-rep}
\big[\forall j, 1 \leq j \leq \k: f_j(u) = f_j(v)\big] \Rightarrow u \equiv v
\end{align}
\noindent
where $f_j$ is a feature.
Eq.~\ref{eq:node-equiv-feature-rep} is strict since two nodes belong to the same type 
iff they have identical feature vectors. 
However, Figure~\ref{fig:attr-tri-2stars-logbinning} captures a more relaxed notion of structural equivalance called structural similarity~\cite{rossi2015-tkde}.
This result is surprising since no embedding has been learned yet, only the mapping into the intermediate representation (types).
Hence, it validates the intermediate representation of mapping nodes to types while also demonstrating the effectiveness of it for grouping structurally similar nodes.

\begin{figure}[b!]
\centering
{\includegraphics[width=0.94\linewidth]{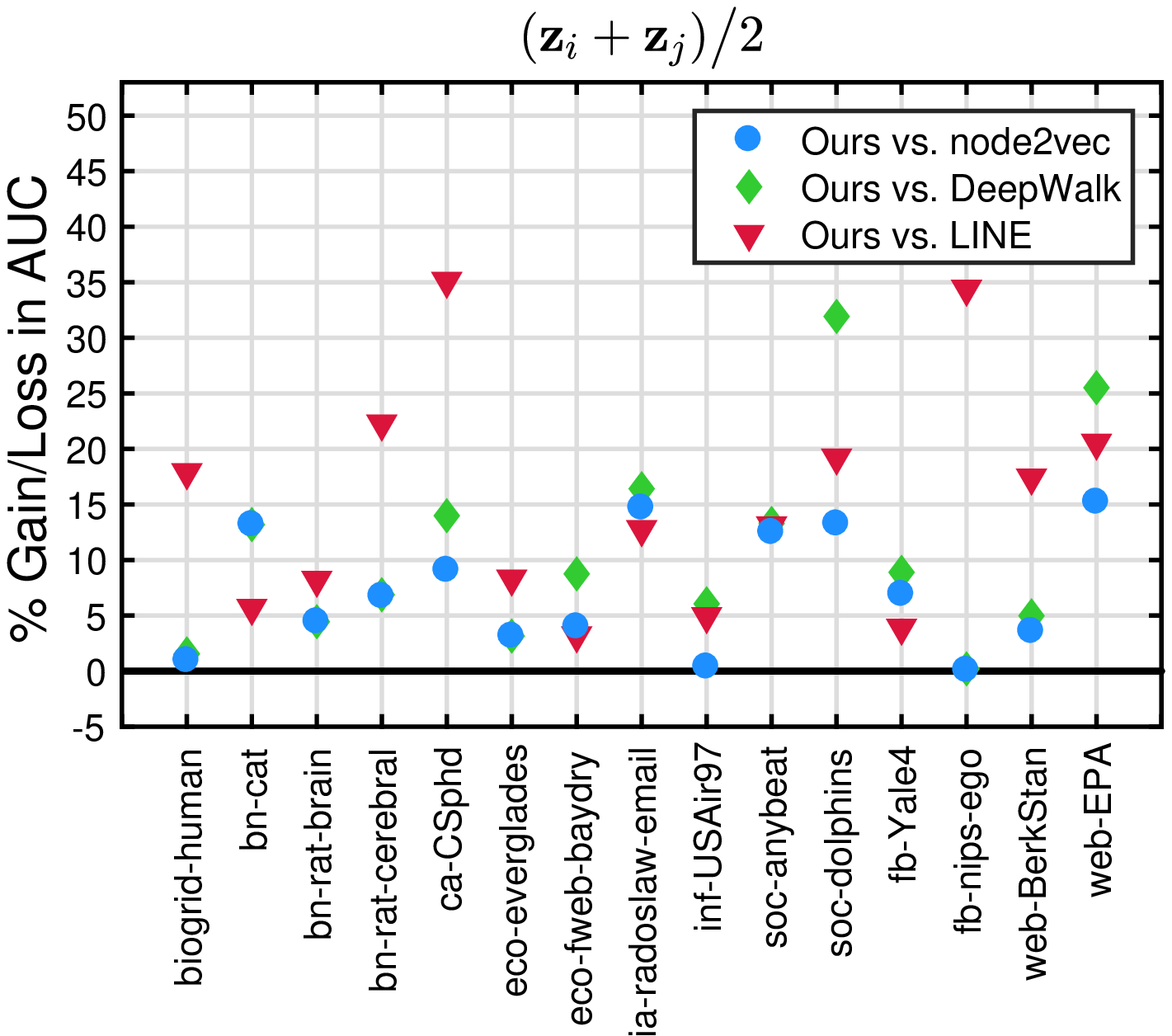}} 
\vspace{-2mm}
\caption{
AUC gain over other methods for link prediction bootstrapped using $(\vz_i + \vz_j)/2$.
The average gain over all methods and across all graphs is $10.5\%$.
}
\label{fig:link-pred-perc-gain-auc-score-mean-op-rm}
\vspace{-2mm}
\end{figure}

\subsection{Comparison}
\label{sec:exp-comparison}
This section compares the proposed approach to other embedding methods for link prediction.
Given a partially observed graph $G$ with a fraction of missing edges, the link prediction task is to predict these missing edges.
We generate a labeled dataset of edges as done in~\cite{node2vec}.
Positive examples are obtained by removing $50\%$ of edges randomly, whereas \emph{negative examples} are generated by randomly sampling an equal number of node pairs that are not connected with an edge, \ie, each node pair $(i,j) \not\in E$. 
For each method, we learn features using the remaining graph that consists of only positive examples.
Using the feature representations from each method, we then learn a model to predict whether a given edge in the test set exists in $E$ or not.
Notice that node embedding methods such as $\deepwalk$ and node2vec require that each node in $G$ appear in at least one edge in the training graph (\ie, the graph remains connected), otherwise these methods are unable to derive features for such nodes.
This is a significant limitation that prohibits their use in many real-world applications.

\begin{figure}[h!]
\centering
{\includegraphics[width=0.94\linewidth]{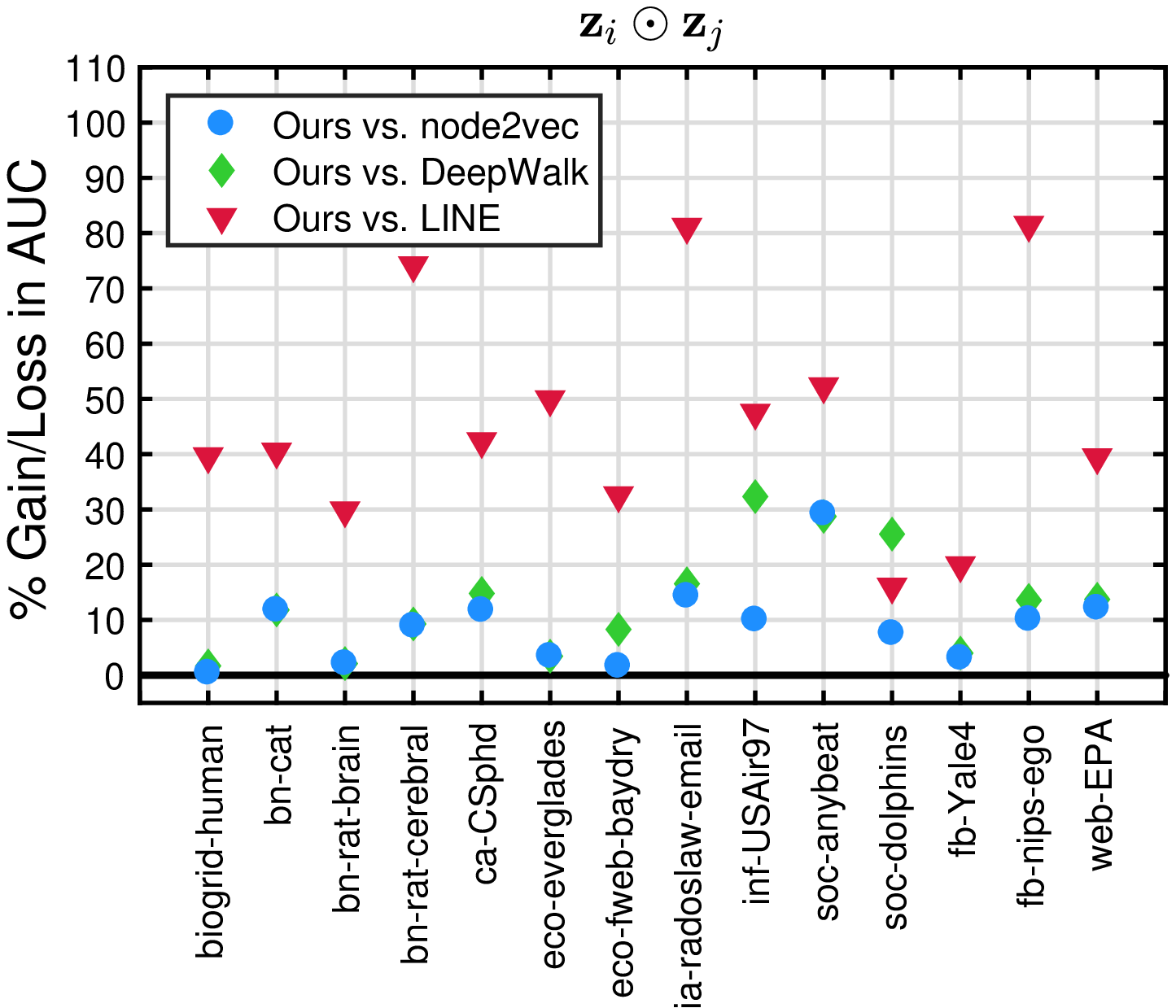}}
\vspace{-2mm}
\caption{
AUC gain of our approach over the other methods for link prediction bootstrapped using Hadamard.
The average gain over all methods and across all graphs is $21.7\%$.
}
\label{fig:link-pred-perc-gain-auc-score-prod-op-rm}
\vspace{-0mm}
\end{figure}

For comparison, we use the same set of binary operators~\cite{node2vec} 
to construct features for the edges \emph{indirectly} using the learned node representations. 
Moreover, the AUC scores from $\ourMethod$ are all significantly better than the other methods at $p<0.01$.
We summarize the gain/loss in predictive performance over the other methods 
in Figure~\ref{fig:link-pred-perc-gain-auc-score-mean-op-rm} and~\ref{fig:link-pred-perc-gain-auc-score-prod-op-rm}.
In all cases, $\ourMethod$ achieves better predictive performance over the other methods 
across a wide variety of graphs with different characteristics.
Overall, the mean and product binary operators give the best results with an average gain in predictive performance (over all graphs) of $10.5\%$ and $21.7\%$, respectively.

\begin{figure*}[h!]
\centering
\includegraphics[width=0.95\linewidth]{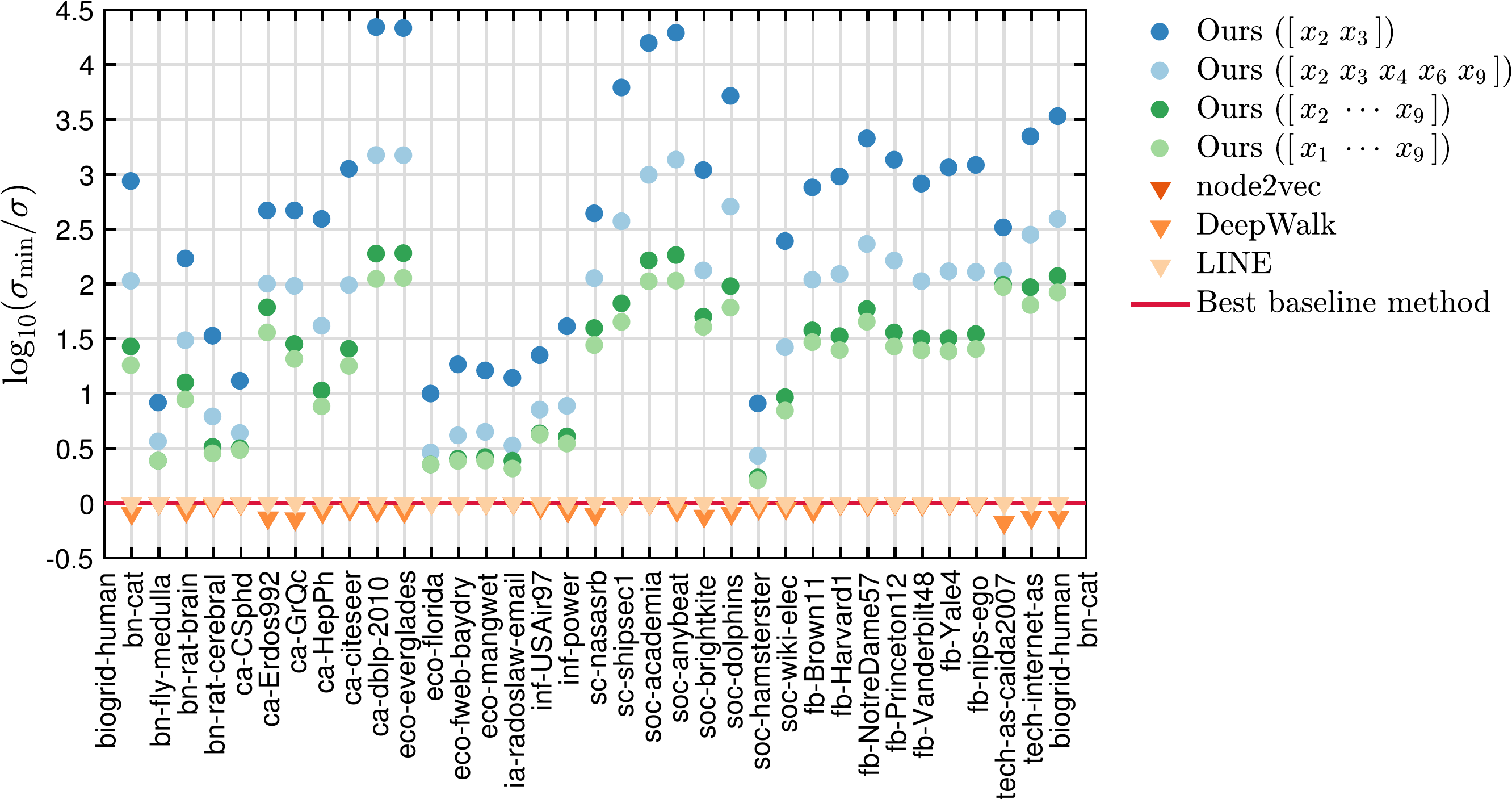} 
\vspace{-3mm}
\caption{
The space requirements of the proposed approach are orders of magnitude less than existing methods.
Note $\log_{10}(\sigma_{\min} / \sigma)$ (y-axis) is the relative ``gain/loss'' (in space) over the best baseline method where $\sigma$ is the size (bytes) of an embedding and $\sigma_{\min}$ is the embedding with the smallest size among the baseline methods (node2vec, DeepWalk, LINE); 
and $[\,\x_2 \; \x_3\,], \cdots, [\,\x_1\; \cdots \; \x_9\,]$ is the attribute sets in Eq.~\eqref{eq:space-1-2stars-triangles}-\eqref{eq:space-4-all} used as input to $\phi$ for mapping nodes to types.
These represent variations of our method.
The baseline method with the minimum space ($\sigma_{\min}$) is shown at $0$ and thus methods that use more space lie below the horizontal line.
}
\label{fig:space-efficient-embedding-reduction-gain}
\end{figure*}

\subsection{Space-efficient Embeddings}
\label{sec:exp-space-efficiency}
We now investigate the space-efficiency of the learned embeddings from the proposed framework and intermediate representation.
Observe that any embedding method that implements the proposed attributed random walk framework (and intermediate representation) learns an embedding for each distinct node type $\y \in \Y$.
In the worst case, an embedding is learned for each of the $\n$ nodes in the graph and we recover the original method as a special case.
In general, the best embedding most often lies between such extremes and therefore the embedding learned from a method implementing the framework is often orders of magnitude smaller in size since $\m \ll \n$ where $\m=|\Y|$ and $\n=|V|$.

Given an attribute vector $\vx$ of graphlet counts (Figure~\ref{fig:graphlet-attributes}) for an arbitrary node in $G$, we derive embeddings using each of the following:
{\fontsize{9}{10}\selectfont
\begin{align}
&\phi(\vx_i=[\,\x_2 \; \x_3\,]), \;\; \text{for } i=1,...,\n \label{eq:space-1-2stars-triangles}\\
&\phi(\vx_i=[\,\x_2 \; \x_3\; \x_4\; \x_6\; \x_9\,]), \;\; \text{for } i=1,...,\n \label{eq:space-2-2stars-triangles-x4-x6-4cliques} \\ 
&\phi(\vx_i=[\, \x_2 \; \x_3\; \x_4\; \x_5\, \cdots\, \x_9\,]), \;\; \text{for } i=1,...,\n
\label{eq:space-3-all-but-degree} \\
&\phi(\vx_i=[\,\x_1\; \x_2 \; \x_3\; \x_4\; \x_5\, \cdots\, \x_9\,]), \;\; \text{for } i=1,...,\n
\label{eq:space-4-all} 
\end{align}
}\normalsize
\noindent
where $\phi(\cdot)$ is a function that maps $\vx_i$ to a type $\y \in \Y$.
In these experiments, we use logarithmic binning (applied to each $\n$-dimensional graphlet feature) with $\alpha=0.5$ and use $\phi$ defined as the concatenation of the logarithmically binned attribute values.
Embeddings are learned using the different subsets of attributes in Eq.~\eqref{eq:space-1-2stars-triangles}-\eqref{eq:space-4-all}.
For instance, Eq.~\eqref{eq:space-1-2stars-triangles} indicates that node types are derived using the (logarithmic binned) number of 2-stars $\x_2$ and triangles $\x_3$ that a given node participates (Figure~\ref{fig:graphlet-attributes}).
To evaluate the space-efficiency of the various methods, we measure the space (in bytes) 
required to store the embedding learned by each method.
In Figure~\ref{fig:space-efficient-embedding-reduction-gain}, we summarize the reduction in space from our approach compared to the other methods.
In all cases, the embeddings learned from our approach require significantly less space and thus more space-efficient.
Specifically, the embeddings from our approach require on average $853$ times less space than the best method averaged across all graphs.
In addition, Table~\ref{table:space-embedding-stats-num-words} provides the number of types derived when using the various attribute combinations.

\begin{table}[t!]
\vspace{-4mm}
\centering
\setlength{\tabcolsep}{5.5pt}
\setlength{\tabcolsep}{6.0pt}
\ra{1.1}
\caption{
Comparing the number of unique types.
}
\vspace{1mm}
\label{table:space-embedding-stats-num-words}
\small
\scriptsize
\footnotesize
\fontsize{8}{9}\selectfont
\begin{tabularx}{1.0\linewidth}{
r H  
r
X
XXX 
HH 
HHHH
@{}
}
\toprule

&&
&
\multicolumn{4}{c}{$\m$}
&&
\\
\cmidrule(rl){4-8} 

\textsc{Graph} && 
$\n$ & 
Eq.~\eqref{eq:space-1-2stars-triangles} &
Eq.~\eqref{eq:space-2-2stars-triangles-x4-x6-4cliques} &
Eq.~\eqref{eq:space-3-all-but-degree} &
Eq.~\eqref{eq:space-4-all} &
\\
\midrule

\dataName{biogrid\text{--}human}  && 
9527 & 9 & 73 & 290 & 429 & 
\\

\dataName{ca\text{--}citeseer}  && 
227320 & 9 & 132 & 1049 & 1783 & 
\\

\dataName{ca\text{--}dblp\text{--}2010}  && 
226413 & 9 & 130 & 1018 & 1706 & 
\\

\dataName{fb\text{--}Harvard1}  && 
15126 & 7 & 64 & 253 & 328 & 
\\

\dataName{fb\text{--}NotreDame57}  && 
12155 & 9 & 75 & 340 & 454 & 
\\

\dataName{sc\text{--}shipsec1}  && 
140385 & 9 & 144 & 866 & 1346 & 
\\

\dataName{soc\text{--}academia}  && 
200169 & 9 & 129 & 954 & 1634 & 
\\

\dataName{soc\text{--}brightkite}  && 
56739 & 9 & 92 & 489 & 768 & 
\\

\dataName{tech\text{--}internet\text{--}as}  && 
40164 & 9 & 78 & 259 & 365 & 
\\

\bottomrule
\end{tabularx}
\end{table}

\section{Related Work} 
\label{sec:related-work}
\noindent

\noindent
Recent embedding techniques for graphs
have largely been based on the popular skip-gram model~\cite{word2vec,cheng2006n} originally introduced for learning vector representations of words in the natural language processing (NLP) domain.
In particular, DeepWalk~\cite{deepwalk} used this approach to embed the nodes such that the co-occurrence frequencies of pairs in short random walks are preserved. 
More recently, node2vec~\cite{node2vec} introduced hyperparameters to DeepWalk that tune the depth and breadth of the random walks.
These approaches are becoming increasingly popular and have been shown to outperform a number of existing methods.
These methods are all based on simple random walks and thus are well-suited for generalization using the \emph{attributed random walk framework}.

On the other hand, methods such as SkipGraph \cite{skipgraph} make use of simple random walks to learn embeddings for entire graphs (as opposed to individual nodes).
These methods can be used for graph-level tasks such as graph classification and clustering. 
Since these methods are still based on simple random walks, it is straightforward to generalize them using our proposed framework.

While most work has focused on transductive (within-network) learning, there has been some recent work on graph-based inductive approaches~\cite{Planetoid,deepGL,GraphSage}.
Yang~\etal proposed an inductive approach called Planetoid~\cite{Planetoid}.
However, Planetoid is an embedding-based approach for semi-supervised learning and does not use any structural features.
Rossi~\etal proposed an inductive approach for (attributed) networks called DeepGL that learns (inductive) relational functions representing compositions of one or more operators applied to an initial set of 
graph features~\cite{deepGL}.
More recently, Hamilton~\etal~\cite{GraphSage} proposed a similar approach that also aggregates features from node neighborhoods.
However, these approaches are not based on random-walks.

Heterogeneous networks~\cite{shi2014hetesim} have also been recently considered~\cite{chang2015heterogeneous,dong2017metapath2vec} as well as attributed networks
~\cite{huang2017label,huang2017accelerated}.
In particular, 
Huang~\etal proposed an approach for attributed networks with labels~\cite{huang2017label} whereas Yang~\etal used text features to learn node representations~\cite{yang2015network}.
Liang~\etal proposed a semi-supervised approach for networks with outliers~\cite{liang2017seano}. 
Bojchevski~\etal proposed an unsupervised rank-based approach~\cite{bojchevski2017deep}.
More recently, Coley~\etal introduced a convolutional approach for attributed molecular graphs that learns graph embeddings~\cite{coley2017convolutional} as opposed to node embeddings.
However, most of these approaches are not inductive nor space-efficient.

Our work is also related to uniform and non-uniform random walks on graphs~\cite{lovasz1993random,chung2007random}.
Random walks are at the heart of many important applications such as ranking~\cite{page1998pagerank}, community detection~\cite{ng2002spectral,pons2006computing}, 
recommendation~\cite{bogers2010movie}, 
link prediction~\cite{liu2010link}, 
influence modeling~\cite{java2006modeling}, search engines~\cite{lassez:latentlinks}, 
image segmentation~\cite{grady2006random}, routing in wireless sensor networks~\cite{servetto2002constrained}, and time-series forecasting~\cite{rossi2012dpr-dynamical}.
These applications and techniques may also benefit from the notion of attributed random walks.

\section{Conclusion}
\label{sec:conc}
To make existing methods more generally applicable, this work proposed a flexible framework based on the notion of attributed random walks.
The framework serves as a basis for generalizing existing techniques (that are based on random walks) for use with attributed graphs, unseen nodes, graph-based transfer learning tasks, and allowing significantly larger graphs due to the inherent space-efficiency of the approach.
Instead of learning individual embeddings for each node, our approach learns embeddings for each type based on functions that map feature vectors to types.
This allows for both inductive and transductive learning.
Furthermore, the framework was shown to have the following desired properties: space-efficient, accurate, inductive, and able to support graphs with attributes (if available).
Finally, the approach is guaranteed to perform at least as well as the original methods since they are recovered as a special case in the framework.

\balance
\setlength{\bibsep}{4pt}
\fontsize{9.pt}{10.pt}\selectfont 
\bibliographystyle{abbrvnat}

\bibliography{paper}

\end{document}